# DISCOVERING STOCK PRICE PREDICTION RULES OF BOMBAY STOCK EXCHANGE USING ROUGH FUZZY MULTI LAYER PERCEPTION NETWORKS

Arindam Chaudhuri[1, $]    Kajal De[2]    Dipak Chatterjee[3]

**ABSTRACT**

In India financial markets have existed for many years. A functionally accented, diverse, efficient and flexible financial system is vital to the national objective of creating a market-driven, productive and competitive economy. Today markets of varying maturity exist in equity, debt, commodities and foreign exchange. Of the 25 stock markets in the country, the most important is Bombay Stock Exchange. In this work we attempt to generate prediction rules scheme for stock price movement at Bombay Stock Exchange using an important Soft Computing paradigm viz., Rough Fuzzy Multi Layer Perception. The use of Computational Intelligence Systems such as Neural Networks, Fuzzy Sets, Genetic Algorithms, etc. for Stock Market Predictions has been widely established. The process is to extract knowledge in the form of rules from daily stock movements. These rules can then be used to guide investors. To increase the efficiency of the prediction process, Rough Sets is used to discretize the data. The methodology uses a Genetic Algorithm to obtain a structured network suitable for both classification and rule extraction. The modular concept, based on divide and conquer strategy, provides accelerated training and a compact network suitable for generating a minimum number of rules with high certainty values. The concept of variable mutation operator is introduced for preserving the localized structure of the constituting Knowledge Based sub-networks, while they are integrated and evolved. Rough Set Dependency Rules are generated directly from the real valued attribute table containing Fuzzy membership values. The paradigm is thus used to develop a rule extraction algorithm. The extracted rules are compared with some of the related rule extraction techniques on the basis of some quantitative performance indices. The proposed methodology extracts rules which are less in number, are accurate, have high certainty factor and have low confusion with less computation time. To the best of our knowledge, this is the first Rough Fuzzy Multi Layer Perception method for discovering stock prediction rules of Bombay Stock Exchange.

==================================================================

**1: Lecturer (Mathematics & Computer Science), Meghnad Saha Institute of Technology, Kolkata, India. Email: arindam_chau@yahoo.co.in;** [$]**Corresponding Author**

**2: Professor in Mathematics, School of Science, Netaji Subhas Open University, Kolkata, India. Email: kajalde@rediffmail.com**

**3: Distinguished Professor, Department of Mathematics, St. Xavier's College, Kolkata, India.**

# 1. INTRODUCTION

The complexity and difficulty of predicting stock prices, with a reasonable level of precision, on one hand, and the emergence of Computational Intelligence techniques [7] such as Neural Networks, Fuzzy Sets, Evolutionary Algorithms, Rough Sets etc., as alternative techniques to the conventional Statistical Regression and Bayesian Models with better performance, on the other hand, have paved the road for the increased usage of these techniques in various areas of Finance [14], [28], [36]. Including in these areas is the utilization of the Genetic Algorithms and Genetic Programming [15], for Portfolio Optimization [5], Stock Selection using Neural Networks [21] and predicting the S&P 100 index using RS [31]. The stock and future traders have come to rely upon various types of Intelligent Systems to make the trading decisions [1], [18]. Several Information Systems have been developed in the recent years for modeling Expertise, Decision Support and complicated Automation tasks [19], [22].

In recent years since its inception, Rough Sets have gained momentum and has been widely used as viable Intelligent Knowledge Discovery techniques in many applications including Finance and Investment areas. For instance, building Trading Systems using the Rough Set Model was studied by several researchers. Ziarko et al. [37], Golan and Edwards [12], [11] applied the Rough Set Model to discover strong Trading Rules, which reflect highly repetitive patterns in data, from historical database of the Toronto Stock Exchange. A detailed review of applications of Rough Set in the financial domain can been found in [34].

The major prima face of this Paper is to present a generic Stock Price Prediction Model using a Modular Evolutionary Approach for designing a Hybrid Connectionist System in Soft Computing framework for both Classification and Rule Generation. The basic building block used is the Rough Fuzzy Multi Layer Perception (RFMLP) Networks [4]. The model is expected to extract knowledge in the form of rules from daily stock movements of the Bombay Stock Exchange (BSE) that would guide investors whether to buy, sell or hold a stock. To increase the efficiency of the Prediction process, Rough Set with Boolean Reasoning (RSBR) Discretization Algorithm is used to discretize the data. The original classification includes splitting the tasks into several subtasks and a number of RFMLP Networks are obtained for each subtask. The sub-network modules are integrated in a particular manner so as to preserve the crude domain knowledge which was encoded in them using Rough Sets. The pool of integrated networks is then evolved using a Genetic Algorithms with a restricted adaptive or variable mutation operator that utilizes the domain knowledge to accelerate training and preserves the localized rule structure as potential solutions. The parameters for input and output fuzzy membership functions of the network are also tuned using Genetic Algorithms together with the link weights. The existing procedure has been modified for the generation of Rough Set Dependency Rules for handling directly the real valued attribute table containing fuzzy membership values. This helps in preserving all the class representative points in the Dependency Rules by adaptively applying a threshold that automatically takes care of the shape of membership functions. In this Knowledge Based Network design all possible inference rules contribute to the final solution. The use of Genetic Algorithms in this context is beneficial for modeling Multimodal distributions, since all major representatives in the population are given fair chance during Network synthesis. Next the Rule Extraction Algorithm is presented. The performance of the generated rules is evaluated quantitatively. Two new measures are accordingly defined indicating the *certainty* and *confusion* in the decision. These new indices are used along with some existing measures to evaluate the quality of the rules. A quantitative comparison of the Rule Extraction Algorithm is made with some existing ones like the Subset Method [9], *M of N* Method [35], X2R Method [20], etc. on datasets of the BSE.

The organization of this Paper is as follows. In section 2, a brief discussion on the Indian Financial Markets is given. In the next two sections discusses the Rough Sets and Fuzzy Multi Layer Perception

Networks respectively. Section 5 presents the RFMLP Prediction Model for generating the Stock Price Prediction Rules of BSE alongwith the relevant design details of the Modular Evolutionary Algorithms. The Quantitative Performance Measures are presented in section 6. In section 7, the Simulation Results alongwith discussions are given. Finally, in section 9 Conclusions are given.

## 2. INDIAN FINANCIAL MARKETS

This section gives a brief overview of the Indian Financial Markets. Organized financial markets have existed in India for more than a century. Today, markets of varying maturity exist in equity, debt, commodities and foreign exchange. There are 25 stock markets all over the country, the most important of which are the Bombay Stock Exchange (BSE) and the National Stock Exchange (NSE). The rupee has been convertible on the current account since 1992. India's total market capitalization touched Rs 644.67 billion, with an average daily turnover of Rs 2,384 million, in December 1995. India's market capitalization was the 6th highest among the emerging markets. The number of companies listed on the BSE at the end of December 1994 was 4,702. This was more than the aggregate total of companies listed in 9 emerging markets including Malaysia, South Africa, Mexico, Taiwan, Korea, Philippines, Thailand, Brazil and Chile. The number of companies was also more than that in developed markets of Japan, United Kingdom, Germany, France, Australia, Switzerland, Canada and Hong Kong. India's capital market features a wide variety of capital market instruments.

The Capital markets remained subdued through most of 1995-96 and the bear phase which began in October 1994, continued through most part of 1995-96. There was a slow down in Foreign Institutional Investors inflow and domestic liquidity conditions were relatively tight. Between April 1995 to December 1995, the value of primary issues was marginally higher than the corresponding period the previous year, despite a downtrend in stock prices and low turnover in stock exchanges. The process of reforms in the capital markets, including the money markets, was further strengthened. Securities and Exchange Board of India (SEBI) was empowered to regulate all market intermediaries. An Ordinance to establish depositories was announced, thus addressing one of the major lacunae in the system. The NSE expanded rapidly, providing an incentive to other stock exchanges to accelerate computerization. India Financial Market promotes the savings of the economy, providing an effective channel for transmitting the financial policies. It is a well-developed, competitive, efficient and integrated financial sector. There are large numbers of buyers and sellers of the financial product, the prices are fixed by the market forces of demand and supply within the Indian Financial Market. The other markets of the economy assist the functioning of the financial market in India.

The Financial Market in India focuses on the important features viz., (i) Real-time India Financial Indices – BSE 30 Index, Sector Indexes, Stock Quotes, Sensex Charts, Bond prices, Foreign Exchange, Rupee and Dollar Chart; (ii) Indian Financial Market News and Stock News – BSE, BSE Sensex 30 closing index, S and P CNX - Nifty NSE, stock quotes, company information, issues on market capitalization, corporate earning statements, Indian Business Directory; (iii) Fixed Income – Corporate Bond Prices, Corporate Debt details and Debt trading activities, Interest Rates, Money Market, Government Securities, Public Sector Debt and External Debt Service; (iv) Foreign Investment – Foreign Debt Database composed by BIS, IMF, OECD and World Bank, Investments in India and Abroad; (v) Global Equity Indexes – Dow Jones Global indexes, Morgan Stanley Equity Indexes; (vi) Currency Indexes – FX and Gold Chart Plotter, J. P. Morgan Currency Indexes; (vii) National and Global Market Relations, Mutual Funds, Insurance, Loans, Forex etc. A clear insight with information on the Indian Financial Market will thus be the most beneficial for the investors and the marketers of both India and the foreign countries.

# 3. ROUGH SETS

This section presents some preliminaries of Rough Sets which are relevant to this work. An *information system* is defined by a pair $S = <U, A>$, where $U$ is a nonempty finite set called the *universe* and $A$ is a nonempty finite set of *attributes*. An attribute $a$ can be regarded as a function from the domain $U$ to some value set $V_a$. A *decision system* is any *information system* of the form $A = (U, A \cup \{d\})$, where $d \notin A$ is the *decision attribute*. The elements of $A$ are called *conditional attributes*. An *information system* can be represented as an *attribute-value table*, in which rows are labeled by objects of the *universe* and columns by the *attributes*. Similarly, a decision system can be represented by a *decision table*. With every subset of attributes $B \subseteq A$, an equivalence relation $I_B$ can easily be associated on $U$, $I_B = \{(x, y) \in U: \forall\, a \in B, a(x) = a(y)\}$. Then, $I_B = \bigcap_{a \in B} I_a$. If $X \subseteq U$, the sets $\{x \in U: [x]_B \subseteq X\}$ and $\{x \in U: [x]_B \cap X = \emptyset\}$, where, $[x]_B$ denotes the equivalence class of the object $x \in U$ relative to $I_B$, are called the *B-lower* and *B-upper approximation* of $X$ in $S$ and denoted by $\underline{B}X, \overline{B}X$ respectively. $X \subseteq U$ is *B-exact* or *B-definable* in $S$ if $\underline{B}X = \overline{B}X$. It may be observed that $\underline{B}X$ is the greatest B-definable set contained in X and $\overline{B}X$ is the smallest B-definable set containing X.

Now we define the notions relevant to knowledge reduction. The aim is to obtain irreducible but essential parts of the knowledge encoded by the given *information system*, which constitutes the *reducts* of the system. This in effect reduces to looking for *maximal* sets of attributes taken from the initial set $A$ which induce the same partition on the domain as $A$. In other words, the essence of the information remains intact and superfluous attributes are removed. Reducts have already been characterized by the *discernibility matrices* and *discernibility functions*. Consider $U = \{x_1,\ldots\ldots,x_n\}$ and $A = \{a_1,\ldots\ldots,a_m\}$ in the *information system* $S = <U, A>$. By the *discernibility matrix* M($S$), of $S$ is meant an $n \times n$ matrix such that

$$c_{ij} = \{a \in A : a(x_i) \neq a(x_j)\} \quad (1)$$

A discernibility function $f_S$ is a function of $m$ Boolean variables $\overline{a}_1,\ldots\ldots,\overline{a}_m$ corresponding to the attributes $a_1,\ldots\ldots,a_m$ respectively and is defined as follows:

$$f_S(\overline{a}_1,\ldots\ldots,\overline{a}_m) = \Lambda\{\vee(c_{ij}) : 1 \leq i, j \leq n, j < i, c_{ij} \neq \Phi\} \quad (2)$$

where, $\vee(c_{ij})$ is the disjunction of all variables $\overline{a}$ with $a \in c_{ij}$. It is observed that $\{a_{i_1},\ldots\ldots,a_{i_p}\}$ is a reduct in $S$ if and only if $a_{i_1} \wedge \ldots\ldots \wedge a_{i_p}$ is a prime implicant of $f_S$.

# 4. FUZZY MULTI LAYER PERCEPTION NETWORKS

The Fuzzy Multi Layer Perception Networks [24] are described briefly in this section. Based on the definitions of the Rough Sets provided in the previous section, they are incorporated in the Fuzzy Multi Layer Perception Networks such that RFMLP Networks are evolved as discussed in the next section. This methodology is utilized for extracting the Dependency rules using the knowledge encoding algorithm for mapping the rules to the parameters of the Fuzzy Multi Layer Perception Networks. The Neural Network Model chosen for the data classification is a Multi Layer Perception Network. A Multi Layer Perception Network consists of a group of nodes arranged in layers. Each node in a layer is connected to all nodes in the next layer by links which have weight associated with them. The input layer contains nodes that represent the input features of the classification problem. A real valued feature is represented by a single node, whereas a discrete feature with $n$ distinct values is represented by $n$ input nodes. The classification strength of the Multi Layer Perception Network is enhanced by incorporating Rough Sets and Fuzzy Sets in the Network which results in the development of the RFMLP Model. It is an important Soft Computing paradigm for Pattern Classification. The Neural Network acts as efficient connectionist between the two. In this hybridization, Fuzzy Sets help in handling linguistic input information and ambiguity in output decision, while Rough Sets extract the domain knowledge for determining the network parameters.

The Fuzzy Multi Layer Perception Model incorporates fuzziness at the input and output levels of the Multi Layer Perception and is capable of handling exact or numerical and inexact or linguistic forms of input data. Any input feature value is described in terms of some combination of membership values in the linguistic property sets *low* (L), *medium* (M) and *high* (H). Class membership values $\mu$ of patterns are represented at the output layer of the Fuzzy Multi Layer Perception. During training, the weights are updated by back propagation errors with respect to these membership values such that the contribution of uncertain vectors is automatically reduced. A four layered feed forward Multi Layer Perception is used here. The output of a neuron in any layer $h$ other than the input layer ($h = 0$), is given as

$$y_j^{(h)} = \frac{1}{1 + \exp(-\sum_i y_i^{(h-1)} w_{ji}^{(h-1)})} \quad (3)$$

where, $y_i^{(h-1)}$ is the state of the $i^{th}$ neuron in the preceding $(h-1)^{th}$ layer and $w_{ji}^{(h-1)}$ is the weight of the connection from the $i^{th}$ neuron in layer $(h-1)$ to the $j^{th}$ neuron in layer $h$. For nodes in the input layer, $y_j^0$ corresponds to the $j^{th}$ component of the input vector. Further it is to be noted that $x_j^{(h)} = \sum_i y_i^{(h-1)} w_{ji}^{(h-1)}$.

## 4.1 INPUT VECTOR

An *n*-dimensional pattern $F_i = [F_{i1}, F_{i2}, \ldots, F_{in}]$ is represented as a 3*n*-dimensional vector

$$F_i = [\mu_{low(F_{i1})}(F_i), \ldots, \mu_{high(F_{in})}(F_i)] = [y_1^{(0)}, y_2^{(0)}, \ldots, y_{3n}^{(0)}] \quad (4)$$

where, the $\mu$ values indicate the membership functions of the corresponding linguistic $\pi$-sets *low*, *medium*, and *high* along each feature axis and $y_1^{(0)}, y_2^{(0)}, \ldots, y_{3n}^{(0)}$ refer to the activations of the 3*n* neurons in the input layer. When the input feature is exact in nature, $\pi$-Fuzzy sets in the one dimensional form are used with range [0, 1] and are represented as

$$\pi(F_j; c, \lambda) = \begin{cases} 2(1 - \frac{\|F_j - c\|}{\lambda})^2, & \text{for } \lambda/2 \leq \|F_j - c\| \leq \lambda \\ 1 - 2(\frac{\|F_j - c\|}{\lambda})^2, & \text{for } 0 \leq \|F_j - c\| \leq \lambda/2 \\ 0, & \text{otherwise} \end{cases} \quad (5)$$

where, $\lambda > 0$ is the radius of the $\pi$-function with $c$ as the central point.

## 4.2 OUTPUT REPRESENTATION

Let us consider an *l*-class problem domain such that we have *l* nodes in the output layer. Also consider the *n*-dimensional vectors $o_k = [o_{k1}, \ldots, o_{kl}]$ and $v_k = [v_{k1}, \ldots, v_{kl}]$ denote the mean and standard deviation respectively, of the exact training data for the $k^{th}$ class $c_k$. The weighted distance of the training pattern $F_i$ from the $k^{th}$ class $c_k$ is defined as

$$z_{ik} = \sqrt{\sum_{j=1}^{n}[\frac{F_{ij} - o_{kj}}{v_{kj}}]^2} \quad \text{for } k = 1,\ldots\ldots,l \quad (6)$$

where, $F_{ij}$ is the value of the $j^{th}$ component of the $i^{th}$ pattern point. The membership of the $i^{th}$ pattern in class $k$, lying in the range [0, 1] is defined as:

$$\mu_k(F_i) = \frac{1}{1 + (\frac{z_{ik}}{f_d})^{f_e}} \quad (7)$$

where, positive constants $f_d$ and $f_e$ are the denominational and exponential fuzzy generators controlling the amount of fuzziness in the class membership set.

## 5. GENERATION OF THE STOCK PRICE PREDICTION RULES USING ROUGH FUZZY MULTI LAYER PERCEPTION PREDICTION MODEL

In this section we discuss the methodology to discover the Stock Price Prediction Rules of BSE. It consists of the five phases viz., Pre-processing Phase, Analysis Phase, Rule Generating Phase, Classification Phase and Rule Extraction and Prediction Phase. These phases are detailed in the following subsections. For the sake of convenience the Analysis Phase and the Rule Generating Phase are clubbed in one subsection. The implementation of the methodology is done using the MATLAB Software Package and the C Programming language.

### 5.1. PRE PROCESSING PHASE

In this phase, the decision table is created for the Rough Set analysis is created. In this process, a number of data preparation tasks such as data conversion, data cleansing, data completion checks, conditional attribute creation, decision attribute generation, discretization of attributes are performed. Data splitting is performed which created two randomly generated subsets, one subset containing objects for the analysis and the remaining subset containing objects for validation. It must be emphasized that data conversion performed on the initial data must generate a form in which specific Rough Set tools can be applied.

### 5.1.1 DATA COMPLETION AND DISCRETIZATION OF CONTINUOUS VALUED ATTRIBUTES

The real world data often contain missing values. Since Rough Set classification involves mining for rules from the data, objects with missing values in the data set may have undesirable effects on the rules that are constructed. The aim of the data completion procedure is to remove all objects that have one or more missing values. Incomplete data or information systems exist broadly in practical data analysis, and approaches to complete the incomplete information system through various completion methods in the pre-processing stage are normal in the knowledge discovery process. However, these methods may lead to the distortion in the original data and knowledge, and can even render the original data to be unexplored. To overcome these deficiencies inherent in the traditional methods, we used the decomposition approach for incomplete information system (i.e. decision table) as proposed in [26].

Attributes in concept classification and prediction, may have varying importance in the problem domain being considered. Their importance can be pre-assumed using auxiliary knowledge about the problem and expressed by properly chosen weights. However, when using Rough Set approach for concept classification, it avoids any additional information aside from what is included in the information table itself. Basically, the Rough Set approach tries to determine from the available data in the decision table whether all the attributes are of the same strength and, if not, how they differ in respect of the classifier power. Therefore, some strategies for discretization of real value attributes have to be used when we need to apply learning strategies for data classification with real value attributes. It has been shown that the quality of learning algorithm dependents on this strategy [26]. Discretization uses data transformation procedure which involves finding cuts in the data sets that divide the data into intervals. Values lying within an interval are then mapped to the same value. Performing this process leads to reducing the size of the attributes value set and ensures that the rules that are mined are not too specific. For the discretization of continuous-valued attributes, in this work, we consider the Rough Sets with Boolean reasoning (RSBR) algorithm [26]. The main advantage of RSBR is that it combines discretization of real valued attributes and classification. The main steps of the RSBR discretization algorithm are given in the Algorithm 1.

**Algorithm 1: RSBR Discretization Algorithm**

**Input:** Information system table (*S*) with real valued attributes $A_{ij}$ and *n* is the number of intervals for each attribute.
**Output:** Information table (*ST*) with discretized real valued attribute.

**Step1: for** $A_{ij} \in S$ **do**
**Step 2:** Define a set of Boolean variables as follows:
$$B = \{\sum_{i=1}^{N} C_{ai}, \sum_{i=1}^{N} C_{bi}, \sum_{i=1}^{N} C_{ci}, \ldots, \sum_{i=1}^{N} C_{Ni}\} \quad (8)$$
where, $\sum_{i=1}^{N} C_{ai}$ correspond to a set of intervals defined on the variables of attributes *a*
**Step 3: end for**
**Step 4:** Create a new information table $S_{new}$ by using the set of intervals $C_{ai}$

**Step 5:** Find the minimal subset of $C_{ai}$ that discerns all the objects in the decision class *D* using the following formula:
$$Y^u = \wedge\{\Phi(i,j) : d(x_i) \neq d(x_j)\} \quad (9)$$
where, $\Phi(i,j)$ is the number of minimal cuts that must be used to discern two different instances $x_i$ and $x_j$ in the information table.

## 5.2. ANALYSIS AND RULE GENERATING PHASE

The principal task in the method of rule generation is to compute reducts and the corresponding rules with respect to the particular kind of *information system* and the *decision system*. The knowledge encoding is also embedded in this phase. We use relativised versions of the matrices and functions viz., *d-reducts* and *d-discernibility matrices* as the basic tools for computation. The methodology is discussed below.

Let $S = <U, A>$ be a decision table, with *C* and $D = \{d_1, \ldots, d_l\}$ its sets of condition and decision attributes respectively. We divide the decision table $S = <U, A>$ into *l* tables $S_i = <U_i, A_i>$; $i = 1, \ldots, l$, corresponding to the *l* decision attributes $d_1, \ldots, d_l$, where $U = U_1 \cup \ldots \cup U_l$ and $A_i = C \cup \{d_i\}$. Let $\{x_{i1}, \ldots, x_{ip}\}$ be the set of those objects of $U_i$ that occur in $S_i$; $i = 1, \ldots, l$. Now for each $d_i$-reduct $B = \{b_1, \ldots, b_k\}$, a *discernibility matrix* denoted by $M_{d_i}(B)$ from the $d_i$-discernibility matrix is defined as follows:
$$c_{ij} = \{a \in B : a(x_i) \neq a(x_j)\}, \text{ for } i, j = 1, \ldots, n \quad (10)$$

For each object $x_j \in x_{i_1}, \ldots, x_{i_p}$, the discernibility function $f_{d_j}^{x_j}$ is defined as
$$f_{d_j}^{x_j} = \wedge\{\vee(c_{ij}) : 1 \leq i, j \leq n, j < i, c_{ij} \neq \Phi\}, \quad (11)$$

where, $\vee (c_{ij})$ is the disjunction of all members of $c_{ij}$. Then, $f_{d_j}^{x_j}$ is brought to its conjunctive normal form. Thus, a dependency rule $r_i$ is obtained, viz., $P_i \leftarrow d_i$, where $P_i$ is the disjunctive normal form $f_{d_j}^{x_j}$, $j \in i_1,\ldots\ldots,i_p$. The dependency factor $df_i$ for $r_i$ is given by

$$df_i = \frac{card(POS_i(d_i))}{card(U_i)} \quad (12)$$

where, $POS_i(d_i) = \bigcup_{X \in I_{d_i}} l_i(X)$ and $l_i(X)$ is the lower approximation of $X$ with respect to $I_i$. In this case, $df_i = 1$.

### 5.2.1 KNOWLEDGE ENCODING

In the knowledge encoding, consider the feature $F_j$ for class $c_k$ in the $l$-class problem domain. The inputs for the $i^{th}$ representative sample $F_i$ are mapped to the corresponding three-dimensional feature space of $\mu_{low(F_{ij})}(F_i), \mu_{medium(F_{ij})}(F_i), \mu_{high(F_{ij})}(F_i)$ which correspond to the *low*, *medium* and *high* values of the stock price index. Let these be represented by $L_j$, $M_j$, $H_j$ respectively. As the method considers multiple objects in a class, a separate $n_k \times 3n$-dimensional attribute-value decision table is generated for each class $c_k$ where, $n_k$ indicates the number of objects in $c_k$. The absolute distance between each pair of objects is computed along each attribute $L_j$, $M_j$, $H_j$ for all $j$. The equation (10) is modified to directly handle a real-valued attribute table consisting of fuzzy membership values. We define

$$c_{ij} = \{a \in B : | a(x_i) - a(x_j) | > Th\}, \quad (13)$$

for $i, j = 1,\ldots\ldots,n_k$ where, $Th$ is an adaptive threshold. It is to be noted that the adaptivity of this threshold is built in, depending on the inherent shape of the membership function. While designing the initial structure of the RFMLP, the union of the rules of the $l$ classes is considered. The input layer consists of $3n$ attribute values while the output layer is represented by $l$ classes. The hidden layer nodes model the innermost operator in the antecedent part of a rule, which can be either a conjunct or a disjunct. The output layer nodes model the outer level operands, which can again be either a conjunct or a disjunct. For each inner level operator, corresponding to one output class one dependency rule, one hidden node is dedicated. Only those inputs attribute that appear in this conjunct or disjunct are connected to the appropriate hidden node, which, in turn, is connected to the corresponding output node. Each outer level operator is modeled at the output layer by joining the corresponding hidden nodes. It is to be noted that a single attribute involving no inner level operators is directly connected to the appropriate output node via a hidden node to maintain uniformity in rule mapping.

### 5.3. CLASSIFICATION PHASE

The classification phase classifies the rules generated from the previous phase. In the process of classifying the dataset the problem is effectively decomposed into sub-problems as a result of which the sub-problems can be solved with compact networks and efficient combination and training of the networks such that there is gain in terms of training time, network size and accuracy. These are discussed in the following sections.

We use two stages. In the first stage, an $l$ class classification problem is split into $l$ two-class problems. Let there be $l$ sets of sub-networks, with $3n$ inputs and one output node each. Rough Set theoretic concepts are used to encode domain knowledge into each of the sub-networks, using Equations (11), (12), and (13). As explained in the previous section, the number of hidden nodes and connectivity of the knowledge-based sub-networks is automatically determined. Each two-class problem leads to the generation of one or more crude sub-networks, each encoding a particular decision rule. Let each of these constitute pool knowledge-based modules. So, we obtain $m \geq l$ such pools. Each pool $k$ is perturbed to generate a total of $n_k$ sub-networks, such that $n_1 = \ldots\ldots = n_k = \ldots\ldots = n_m$. These pools constitute the initial population of sub-networks, which are then evolved independently using Genetic Algorithms. At the end of the first stage, the modules or sub-networks corresponding to each two-class problem are concatenated to form an initial network for the second stage. The inter module links are initialized to small random values as depicted in the Figure 1 below. A set of such concatenated networks forms the initial population of the Genetic Algorithms. The mutation probability for the inter-module links is now set to a high value, while that of intra-module links is set to a relatively lower value. This restricted mutation helps preserve some of the localized rule structures, already extracted and evolved, as potential solutions. The initial population for the Genetic Algorithms of the entire network is formed from all possible combinations of these individual network modules and random perturbations about them. This ensures that for complex multimodal pattern distributions all the different representative points remain in the population. The algorithm then searches through the reduced space of possible network topologies. The steps are summarized below in the algorithm 2.

**Algorithm 2: Classification Algorithm**

**Input:** An $l$ class classification problem.
**Output:** The pool of RFMLP Neural Networks is evolved using the Genetic Algorithm.

**Step 1:** For each class generate rough set dependency rules using the methodology discussed in Rule Generation Phase.
**Step 2:** Map each of the dependency ru1les to a separate sub-network modules using the methodology discussed in knowledge encoding in Rule Generation Phase.
**Step 3:** Partially evolve each of the sub-networks using the conventional Genetic Algorithms.
**Step 4:** Concatenate the sub-network modules to obtain the complete network. For concatenation the intra-module links are left unchanged while the inter-module links are initialized to low random values. Each of the sub-networks solves a 2-class classification problem, while the concatenated network solves the actual $l$ class problem. Every possible combination of sub-network modules is generated to form a pool of networks.
**Step 5:** The pool of networks is evolved using a modified Genetic Algorithm with an adaptive mutation operator. The mutation probability is set to a low value for the intra-module links and to a high value for the inter-module links.

**Example:** Consider a problem of classifying a two dimensional data into two classes. The input fuzzy function maps the features into a six-dimensional feature space. Let a sample set of rules obtained from the Rough Sets be:

$$c_1 \leftarrow (L_1 \wedge M_2) \vee (H_2 \wedge M_1); c_2 \leftarrow M_2 \vee H_1; c_3 \leftarrow L_3 \vee L_1$$

where, $L_j$, $M_j$, $H_j$ correspond to $\mu_{low(F_j)}$, $\mu_{medium(F_j)}$, $\mu_{high(F_j)}$ respectively which denote the *low*, *medium* and *high* values of the stock price index. For the first phase of the Genetic Algorithms three different pools are formed, using one crude sub-network for class 1 and two crude sub-networks for class 2 respectively. Three partially trained sub-networks result from each of these pools. They are then concatenated to form (1 × 2 = 2) networks. The population for the final phase of the Genetic Algorithms is formed with these networks and perturbations about them. The steps followed in obtaining the final network is illustrated in the Figure 2.

### 5.3.1 EVOLUTIONARY DESIGN

We discuss different features of the Genetic Algorithm [13] with relevance to this algorithm.

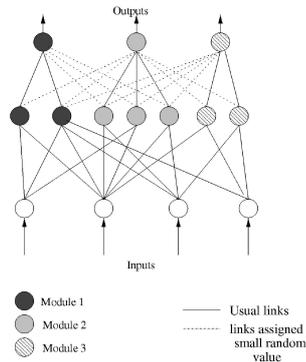

Figure 1: Intra and Inter-module links

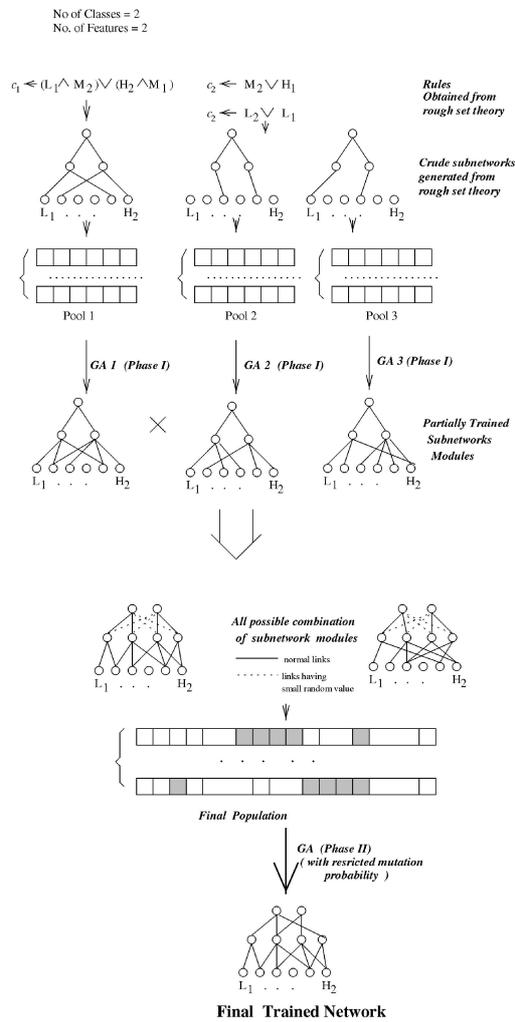

Figure 2: Steps for designing a sample Modular RFMLP

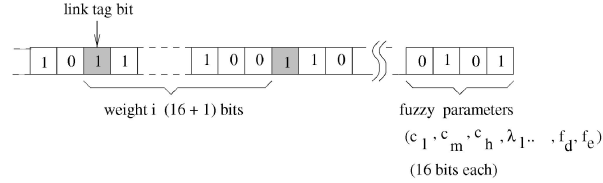

**Figure 3: Chromosomal Representation**

### 5.3.1.1 CHROMOSOMAL REPRESENTATION

The problem variables consist of the weight values and the input/output fuzzy parameters. Each of the weights is encoded into a binary word of 16 bit length, where [000...0] decodes to -128 and [111...1] decodes to 128. An additional bit is assigned to each weight to indicate the presence or absence of the link. The fuzzy parameters tuned are the centers $c$ and radius $\lambda$ for each of the linguistic attributes *low*, *medium*, and *high* of each feature, and the output fuzzy parameters $f_d$ and $f_e$ [24]. These are also coded as 16 bit strings in the range [0, 2]. For the input parameters, [000...0] decodes to 0 and [111...1] decodes to 1.2 times the maximum value attained by the corresponding feature in the training set. The chromosome is obtained by concatenating all the above strings. Sample values of the string length are around 2000 bits for reasonably sized networks.

Initial population is generated by coding the networks obtained by Rough Set based knowledge encoding and by random perturbations about them. A population size of 64 was considered.

### 5.3.1.2 CROSSOVER

It is obvious that due to the large string length, single point crossover would have little effectiveness. Multiple point crossovers are adopted, with the distance between two crossover points being a random variable between eight and 24 bits. This is done to ensure a high probability for only one crossover point occurring within a word encoding a single weight. The crossover probability is fixed at 0.7.

### 5.3.1.3 MUTATION

The search string being very large, the influence of mutation is more on the search compared to crossover. The mutation probability has a spatio-temporal variation. The maximum value of *pmut* is chosen to be 0.4 and the minimum value as 0.01. The mutation probabilities also vary along the encoded string, the bits corresponding to inter-module links being assigned a probability *pmut* i.e., the value of *pmut* at that iteration and intra-module links assigned a probability *pmut* /10. This is done to ensure least alterations in the structure of the individual modules already evolved. Hence, the mutation operator indirectly incorporates the domain knowledge extracted through the Rough Sets.

### 5.3.1.4 CHOICE OF FITNESS FUNCTION

The objective function considered is of the form: $F = \alpha_1 f_1 + \alpha_2 f_2$ (14) where, $f_1 =$ (No. of correctly classified sample in training set) / (Total no. of samples in training set); $f_2 = 1 -$ (No. of links present) / (Total no of links possible). Here, $\alpha_1$ and $\alpha_2$ determine the relative weight of each of the factors. $\alpha_1$ is taken as 0.9 and $\alpha_2$ is taken as 0.1, to give more importance to the classification score compared to the network size in terms of number of links. It is to be noted that we optimize the network connectivity, weights and input or output fuzzy parameters simultaneously. Selection is done by the *roulette wheel* method. The probabilities are calculated on the basis of ranking of the individuals in terms of the objective function, instead of the objective function itself. *Elitism* is incorporated in the selection process to prevent oscillation of the fitness function with generation. The fitness of the best individual of a new generation is compared with that of the current generation. If the latter has a higher value the corresponding individual replaces a randomly selected individual in the new population.

### 5.4. RULE EXTRACTION AND PREDICTION PHASE

This phase extracts the rules from the previous phases and utilizes them to predict the stock price movement. The algorithm for rule extraction considered here is decompositional in nature which is given below in the Algorithm 3.

**Algorithm 3: Decompositional Algorithm**

**Input:** Several RFMLP hybrid Neural Networks.
**Output:** Generates most of the embedded rules over a small number of computational steps.

**Step 1:** Compute the following quantities: $P_{Mean} = Mean$ of all positive weights, $P_{Threshold_1} = Mean$ of all positive weights less than $P_{Mean}$, $P_{Threshold_2} = Mean$ of all weights greater than $P_{Mean}$. Similarly calculate $N_{Threshold_1}, N_{Threshold_2}$ for negative weights.

**Step 2: for** each hidden and output unit

    **for** all weights greater than $P_{Threshold_1}$ search for positive rules only and for all weights less than $N_{Threshold_1}$ search for negated rules only by *Subset* method.

    search for combinations of positive weights above $P_{Threshold_2}$ and negative weights greater than $N_{Threshold_1}$ that exceeds the bias. Similarly search for negative weights less than $N_{Threshold_2}$ and positive weights below $P_{Threshold_1}$ to find out the rules.

**Step 3:** Associate with each rule *j* a confidence factor

$$cf_j = \inf_{j:nodes\,in\,the\,path} \frac{(\sum_i w_{ji} - \theta_j)}{\sum_i w_{ji}} \quad (15)$$

where, $w_{ji}$ is the $i^{th}$ incoming link weight to node *j*.

Since, the training algorithm imposes a structure on the network, resulting in a sparse network having few strong links, the $P_{Threshold}$ and $N_{Threshold_1}$ values are well separated. Hence, the above rule extraction algorithm generates most of the embedded rules over a small number of computational steps. An important consideration is the order of application of rules in a rule base. Since most of the real life patterns are noisy and overlapping, rule bases obtained are often not totally consistent. Hence, multiple rules may fire for a single example. Several existing approaches apply the rules sequentially [9], often leading to degraded performance. The rules extracted by our method have confidence factors associated with them. Therefore, if multiple rules are fired we use the strongest rule having the highest confidence.

Two existing rule extraction algorithms, similar in nature to this algorithm, are the *Subset* method [9] and *M of N* method [35]. The major problem with the *Subset* algorithm is that the cost of finding all subsets grows as the size of the power set of the links to each unit. It requires lengthy, exhaustive searches of size $2^k$ for a hidden or output node with a fan-in of *k* and extracts a large set of rules, up to $\beta_p * (1 + \beta_n)$, where $\beta_p$ and $\beta_n$ are the number of subsets of positively and negatively weighted links respectively. Some of the generated rules may be repetitive, as permutations of rule antecedents are not taken care of automatically. Moreover, there is no guarantee that all useful knowledge embedded in the trained network will be extracted. Additionally, the rule extraction procedure involves a back-propagation step requiring significant computation time. The algorithm has good generalization accuracy, but can have degraded comprehensibility [2]. It is to be noted that the groups of links is considered as equivalence classes, thereby generating a bound on the number of rules rather than establishing a ceiling on the number of antecedents.

## 6. QUANTITATIVE PERFORMANCE MEASURES

In this section, we provide some measures in order to evaluate the performance of the rules. Let $N$ be a $l \times l$ matrix whose $(i, j)^{th}$ element $n_{ij}$ indicate the number of patterns actually belonging to class $i$, but classified as class $j$. The Quantitative Performance Measures are as follows:

1. *Accuracy*: It is the correct classification percentage, provided by the rules on a test set defined as $\frac{n_{i_c}}{n_i} \times 100$ (16) where, $n_i$ is equal to the number of points in class $i$ and $n_{i_c}$ of these points is correctly classified.

2. *User's Accuracy*: If $n_i'$ points are found to be classified into class $i$, [29] then the user's accuracy ($U$) is defined as $U = n_{i_c} / n_i'$ (17). This gives a measure of the confidence that a classifier attributes to a region as belonging to a class. In other words, it denotes the level of purity associated with a region.

3. *Kappa*: The coefficient of agreement called *kappa* [29] measures the relationship of beyond chance agreement to expected disagreement. It uses all the cells in the confusion matrix, not just the diagonal elements. The estimate of kappa, $K$ is the proportion of agreement after chance agreement is removed from consideration. The kappa value for class $i$ ($K_i$) is defined as $K_i = \frac{n.n_{ic} - n_i.n_i'}{n.n_i' - n_i.n_i'}$ (18). The numerator and denominator of overall kappa are obtained by summing the respective numerators and denominators of $K_i$ separately over all classes.

4. *Fidelity*: This represents how closely the rule base approximates the parent neural network model [2]. We measure this as the percentage of the test set for which network and the rule base output agree. It is to be note that the fidelity may or may not be greater than accuracy.

5. *Confusion*: This measure quantifies the goal that the *Confusion should be restricted within minimum number of classes*. This property is helpful in higher level decision making. Let $\hat{n}_{ij}$ be the mean of all $n_{ij}$ for $i \neq j$. Then, we define: $Conf = \frac{Card\{n_{ij} : n_{ij} \geq \hat{n}_{ij}, i \neq j\}}{l}$ (19) for an $l$ class problem. The lower the value of *Conf*, lesser is the number of classes between which confusion is occurs.

6. *Cover*: Ideally the rules extracted should cover all the cluster regions of the pattern space. We use the percentage of examples from a test set for which no rules are fired as a measure of the uncovered region. A rule base having a smaller uncovered region is superior.

7. ***Rule base size*:** It is measured in terms of the number of rules. Lower the value is, the more compact is the rule base.

8. ***Computational complexity*:** Here, we present the CPU time required.

9. ***Certainty*:** By certainty of a rule base, we quantify the confidence of the rules as defined by the certainty factor, *cf* .

# 7. SIMULATION RESULTS

## 7.1 DATA SET

To test and verify the prediction capability of the Genetic Rough Neuro Fuzzy Algorithm viz., RFMLP Model, the daily stock movement of the Bombay Stock Exchange, spanning over a period of 10 years (1999 - 2008) were considered.

## 7.2 RESULTS AND DISCUSSION

Let the RFMLP Model be termed as Model S. Some other models are compared with this model to test its effectiveness. The models considered are as follows:

1. **Model O:** An ordinary Multi Layer Perception trained using back propagation with weight decay.

2. **Model F:** A Fuzzy Multi Layer Perception trained using back propagation [24] with weight decay.

3. **Model R:** A Fuzzy Multi Layer Perception trained using back propagation with weight decay and with initial knowledge encoding using Rough Sets [4].

4. **Model FM:** A modular Fuzzy Multi Layer Perception trained with Genetic Algorithms along with tuning of the fuzzy parameters. Here, the term modular refers to the use of sub networks corresponding to each class that are later concatenated using Genetic Algorithms.

Recognition scores obtained for each of the data by the Model S are presented in the Table 2. It also shows a comparison with other related MLP based classification methods viz., Models O, F, R and FM. In all the cases, 10 percent of the samples are used as training set and the remaining samples are used as test set. 10 such independent runs are performed and the mean value and standard deviation of the classification accuracy which are computed over them are presented in the Table 2. The rule generation phase computes the reducts and the corresponding rules with respect to the Stock Exchange data considered. The Dependency Rules generated using the RS and the encoding scheme for the Stock Exchange dataset along with the Input Fuzzy Parameter Values are given in Table 1. The feature $F_i$ in Table 1, where *F* stands for *low*, *medium*, or *high* denotes a property *F* of the $i^{th}$ feature [24]. The integrated networks contain 18 hidden nodes in a single layer for the Stock Exchange dataset. After combination 96 such networks were obtained. The initial population of the Genetic Algorithms was formed using 64 networks. In the first phase of the Genetic Algorithms, for the Models FM and S, each of the sub networks are partially trained for 10 sweeps. The classification accuracies obtained by the Models are analyzed for statistical significance. Tests of significance are performed for the inequality of means (of accuracies) obtained using the RFMLP Algorithm and compared with the other methods considered. Since both mean pairs and the variance pairs are unknown and different, a generalized version of the *t*-test is appropriate in this context. This problem is the classical Behrens-Fisher problem in hypothesis testing; for which a suitable test statistic is described in [17] and tabled in [3]. The test statistic is of the form:

$$v = \frac{\bar{x}_1 - \bar{x}_2}{\sqrt{\lambda_1 s_1^2 + \lambda_2 s_2^2}} \quad (20)$$

where, $\bar{x}_1$, $\bar{x}_2$ are the means, $s_1$, $s_2$ are the standard deviations, $\lambda_1 = 1/n_1$, $\lambda_2 = 1/n_2$ and $n_1$, $n_2$ are the number of observations. Since, experiments were performed on 10 independent random training sets for all the algorithms, we have $n_1 = n_2 = 10$. The test confidence level considered was 95 percent. In Table 2, we present the mean and standard deviation (SD) of the accuracies. Using the means and standard deviations, the value of the test statistics is computed. If the value exceeds the corresponding tabled value, the means are unequal with statistical significance i.e., the algorithm having higher mean accuracy being significantly superior to the one having lower value.

It is observed from Table 2 that Model S performs the best with the least network size as well as least number of sweeps. For Model R and Model F, the classification performance on test set is marginally better than that of Model S, but with significantly higher number of links and training sweeps required. Comparing Models F and R, we observe that the incorporation of domain knowledge in the latter through RS boosts its performance. Similarly, using the modular approach with the Genetic Algorithm (Model FM) improves the efficiency of Model F. Since Model S encompasses the principle of both Models R and FM, it results in the least redundant yet most effective model. The variation of the classification accuracy of the models with iteration is also studied. As expected, Model S is found to have high recognition score at the very beginning of evolutionary training; the next values are attained by Models R and FM, and the lowest being attained by models O and F using back propagation. Model S converges after about 90 iterations of the Genetic Algorithm providing the highest accuracy compared to all the other models. The back propagation based Models require about 2000 – 5000 iterations for convergence. It may be noted that the training algorithm suggested is successful in imposing a structure among the connection weights. It has been observed that the weight values for a Fuzzy Multi Layer Perception trained with back propagation (Model F) are more or less uniformly distributed between the maximum and minimum values. The RFMLP (Model S) has most of its weight values zero while majority of its nonzero weights have a high value. Hence, it can be inferred that the former model results in a dense network with weak links, while the incorporation of Rough Sets, modular concepts and the Genetic Algorithms produces a sparse network with strong links. The latter is suitable for rule extraction. The connectivity (positive weights) of the trained network is shown in the Figure 4 below.

We use the Decompositional Algorithm explained in Rule Extraction and Prediction Phase to extract the rules from the trained network of Model S. These rules are compared with those obtained by the Subset Method [9], *M of N* Method [35], a pedagogical Method X2R [20], and a Decision Tree Based Method C4.5 [27] in terms of the performance measures. The set of rules extracted from the Model S is presented in Table 4 along with their certainty factors (*cf*). The values of the fuzzy parameters of the membership functions L, M, and H are also mentioned. For the stock dataset, we present the fuzzy parameters only for those features that appear in the extracted rules. A comparison of the performance indices of the extracted rules is presented in Table 3.

Since the network obtained using Model S contains fewer links, the generated rules are less in number and they have high *certainty factor*. Accordingly, it possesses relatively higher percentage of uncovered region, though the accuracy did not suffer much. Although the Subset algorithm achieves the highest accuracy, it requires the largest number of rules and computation time. In fact, the accuracy or the computation time of Subset method is marginally better/worse than Model S, while the size of the rule base is significantly less for Model S.

The accuracy achieved by Model S is better than that of *M of N*, X2R, and C4.5. Also, considering *user's accuracy* and *kappa*, the best performance is obtained by Model S. The X2R algorithm requires least computation time but achieves the least accuracy with more rules. The *Conf* index is the minimum for rules extracted by Model S; it also has high *fidelity* viz., 94.22 % for the stock dataset. In a part of the experiment, we also conducted a comparison with Models F, R, and FM for rule extraction. It was observed that the performance degrades substantially for them because these networks are less structured and hence less suitable, as compared to Model S, for rule extraction.

| Dependency Rules: |
|---|
| $c_1 \leftarrow M_2 \vee L_3$ |
| $c_1 \leftarrow M_1 \vee M_2$ |
| $c_2 \leftarrow M_2 \vee M_3 \vee (H_2 \wedge M_2)$ |
| $c_2 \leftarrow M_1 \vee H_2$ |
| $c_3 \leftarrow (L_2 \wedge H_3) \vee (M_2 \wedge H_3)$ |
| $c_3 \leftarrow (L_1 \wedge H_2) \vee (L_1 \wedge M_3)$ |
| $c_4 \leftarrow (L_1 \wedge L_2) \vee (L_1 \wedge L_3) \vee (L_2 \wedge M_3) \vee (L_1 \wedge M_3)$ |
| $c_5 \leftarrow (H_1 \wedge M_2) \vee (M_2 \wedge M_3) \vee (M_1 \wedge M_2) \vee (M_2 \wedge L_3)$ |
| $c_5 \leftarrow (H_1 \wedge M_2) \vee (M_1 \wedge M_2) \vee (H_2 \wedge H_3) \vee (H_2 \wedge L_1)$ |
| $c_5 \leftarrow (L_2 \wedge L_3) \vee (H_3 \wedge M_3) \vee M_1$ |
| $c_6 \leftarrow L_1 \vee M_3 \vee L_2$ |
| $c_6 \leftarrow M_2 \vee H_3$ |
| $c_6 \leftarrow L_2 \vee H_3$ |
| $c_6 \leftarrow M_1 \vee M_3 \vee L_2$ |
| **Fuzzy Parameters:** |
| **Feature 1:** $c_L = 0.346, c_M = 0.469, c_H = 0.619, \lambda_L = 0.117, \lambda_M = 0.169, \lambda_H = 0.136$ |
| **Feature 2:** $c_L = 0.221, c_M = 0.437, c_H = 0.729, \lambda_L = 0.218, \lambda_M = 0.266, \lambda_H = 0.286$ |
| **Feature 3:** $c_L = 0.399, c_M = 0.546, c_H = 0.686, \lambda_L = 0.146, \lambda_M = 0.147, \lambda_H = 0.136$ |

**Table 1: Rough Set Dependency Rules for the BSE Data alongwith the Input Fuzzy Parameter Values**

| Models | Model O | | Model F | | Model R | | Model FM | | Model S | |
|---|---|---|---|---|---|---|---|---|---|---|
| | Train | Test | Train | Test | Train | Test | Train | Test | Train | Test |
| **Accuracy % Mean SD** | 66.9 0.56 | 65.4 0.56 | 86.6 0.46 | 82.7 0.56 | 87.6 0.36 | 87.0 0.26 | 86.4 0.46 | 83.6 0.56 | 88.6 0.26 | 86.4 0.26 |
| **Number of Links** | 136 | | 210 | | 156 | | 124 | | 84 | |
| **Sweeps** | 5600 | | 5600 | | 2000 | | 200 | | 90 | |

**Table 2: Comparative Performance of different Models on the BSE Dataset**

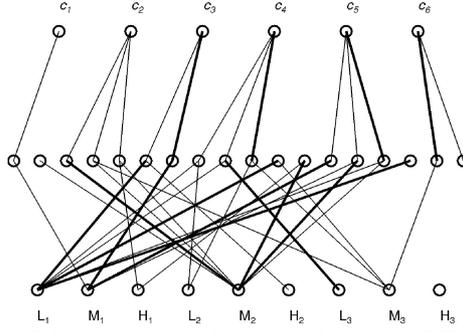

**Figure 4:** Positive connectivity of the network obtained for the BSE Data using Model S. Bold lines indicate weights greater than $PThres_2$, while others indicate values between $PThres_1$ and $PThres_2$.

| Algorithm | Accuracy (%) | Users' Accuracy (%) | Kappa (%) | Uncovered Region (%) | Number of Rules | CPU Time (Sec) | *Conf* |
|---|---|---|---|---|---|---|---|
| **Model S** | 86.04 | 84.37 | 79.19 | 4.10 | 10 | 1.0 | 1.4 |
| **Subset** | 84.00 | 83.72 | 78.29 | 3.89 | 16 | 1.2 | 1.7 |
| **M of N** | 80.00 | 81.04 | 75.69 | 3.20 | 14 | 1.1 | 1.7 |
| **X2R** | 78.00 | 76.89 | 75.36 | 3.72 | 14 | 0.8 | 1.5 |
| **C4.5** | 81.00 | 81.19 | 78.29 | 4.10 | 16 | 0.9 | 1.4 |

**Table 3:** Comparison of the Performance of the Rules extracted by the various Methods on the BSE Data

| **Extracted Rules:** |
|---|
| $c_1 \leftarrow M_1 \vee L_3 \vee M_2$; $cf = 0.856$ |
| $c_1 \leftarrow H_1 \vee M_2$; $cf = 0.766$ |
| $c_2 \leftarrow M_2 \vee M_3$; $cf = 0.829$ |
| $c_2 \leftarrow \neg M_1 \wedge H_1 \wedge L_2 \wedge M_2$; $cf = 0.869$ |
| $c_3 \leftarrow L_1 \vee H_2$; $cf = 0.796$ |
| $c_4 \leftarrow L_1 \wedge L_2 \wedge \neg L_3$; $cf = 0.736$ |
| $c_5 \leftarrow M_2 \wedge H_3$; $cf = 0.896$ |
| $c_5 \leftarrow M_1 \wedge M_2$; $cf = 0.799$ |
| $c_5 \leftarrow H_1 \wedge M_2$; $cf = 0.737$ |
| $c_6 \leftarrow \neg H_2$; $cf = 0.746$ |
| **Fuzzy Parameters:** |
| **Feature 1:** $c_L = 0.360$, $c_M = 0.516$, $c_H = 0.697$ |
| **Feature 1:** $\lambda_L = 0.129$, $\lambda_M = 0.175$, $\lambda_H = 0.186$ |
| **Feature 2:** $c_L = 0.237$, $c_M = 0.446$, $c_H = 0.736$ |
| **Feature 2:** $\lambda_L = 0.219$, $\lambda_M = 0.269$, $\lambda_H = 0.299$ |
| **Feature 3:** $c_L = 0.397$, $c_M = 0.566$, $c_H = 0.687$ |
| **Feature 3:** $\lambda_L = 0.265$, $\lambda_M = 0.219$, $\lambda_H = 0.236$ |

**Table 4:** Rules Extracted from the Trained Networks (Model S) for the BSE Data alongwith the Input Fuzzy Parameter Values

# 8. CONCLUSIONS

In this Paper we presented a methodology for generating stock price prediction rules of BSE using a RFMLP Network. The RFMLP Network was modularly evaluated using Genetic Algorithms for designing a knowledge based network for pattern classification and rule generation. The algorithm involves synthesis of several Multi Layer Perception Modules, each encoding the Rough Set rules for a particular class. These knowledge-based modules are refined using Genetic Algorithms. The genetic operators are implemented in such a way that they help preserve the modular structure already evolved. It is seen that this methodology along with modular network decomposition results in accelerated training and more compact network with comparable classification accuracy, as compared to other hybridizations. The model is used to develop a new rule extraction algorithm. The extracted rules are compared with some of the related rule extraction techniques on the basis of some quantitative performance indices. It is observed that the proposed methodology extracts rules which are less in number, yet accurate and have high certainty factor and low confusion with less computation time. The research has immense potential for application to large scale prediction problems involving knowledge discovery tasks using case-based reasoning particularly related to the mining of classification rules.